\documentclass{article}
\usepackage{spconf,amsmath,graphicx,hyperref,multirow}

\title{CAMEO: Collection of Multilingual Emotional Speech Corpora}
\name{Iwona Christop \qquad Maciej Czajka}
\address{
    Adam Mickiewicz University, ul. Uniwersytetu Poznańskiego 4, 61-614 Poznań, Poland
}

\begin{document}
%
\maketitle
\begin{abstract}
This paper presents CAMEO -- a curated collection of multilingual emotional speech datasets designed to facilitate research in emotion recognition and other speech-related tasks. The main objectives were to ensure easy access to the data, to allow reproducibility of the results, and to provide a standardized benchmark for evaluating speech emotion recognition (SER) systems across different emotional states and languages. The paper describes the dataset selection criteria, the curation and normalization process, and provides performance results for several models. The collection, along with metadata, and a leaderboard, is publicly available via the Hugging Face platform.
\end{abstract}
\section{Introduction}
\label{sec:introduction}

In recent years, speech emotion recognition (SER) has gained attention due to its potential applications in human-computer interaction, mental health monitoring, and customer service. However, developing effective SER systems depends on high-quality datasets that capture the interplay of vocal cues and emotional states.

Despite the growing number of emotional speech corpora, the field still lacks standardization. Inconsistencies are evident in emotion taxonomies, transcription availability, and inclusion of speaker information. The metadata is often scattered across publications rather than linked to audio samples, which limits its usability. Moreover, the access to these datasets is also challenging. Unlike automatic speech recognition (ASR) or text-to-speech (TTS) corpora, which are commonly hosted on platforms like Hugging Face or GitHub, SER datasets are more difficult to locate, and some are shared without proper licensing.

These issues are exacerbated by the absence of standardized tools and benchmarks. Progress toward robust, multilingual SER systems is hindered by the fact that many studies evaluate models on a single dataset or language.

To address these challenges, we introduce \textbf{CAMEO}, a curated, publicly available collection of emotional speech datasets designed for reproducible benchmarking and multilingual evaluation.

The main contributions of this work include:
\begin{itemize}
    \item Release of the \textbf{CAMEO} collection on Hugging Face for transparent and reproducible SER benchmarking\footnote{https://huggingface.co/datasets/amu-cai/CAMEO}.
    \item Systematic curation and normalization of 13 datasets spanning eight languages.
    \item Comparative evaluation of multiple models on the unified corpus using standardized prompts and metrics.
    \item A public leaderboard on Hugging Face for ongoing model comparison\footnote{https://huggingface.co/spaces/amu-cai/cameo-leaderboard}.
\end{itemize}

\section{Related Work}
\label{sec:related-work}

One of the primary limitations of existing emotional speech datasets is their monolingual scope, which hinders the development of multilingual SER systems. Since most corpora cover only one language, the results obtained using them often fail to generalize across languages and cultures, where emotional expression varies.

Several studies have highlighted this issue. For example, Donuk et al.~\cite{article_1214312} reported~66.01\% accuracy on the CREMA-D~\cite{cremad} dataset using convolutional neural networks with particle swarm optimization, while Catania et al.~\cite{10879457} achieved~82.34\% accuracy on the Italian Emozionalmente dataset by fine-tuning wav2vec~2.0. Though promising, such results remain language-specific. Similarly, Mocanu et al.~\cite{MOCANU2023104676} showed that multimodal systems combining audio and video outperform audio-only approaches. On the RAVDESS dataset~\cite{ravdess}, for example, accuracy increased from~76\%~(audio only) to~83\%~(audio and video). On the CREMA-D dataset, accuracy increased from~62\% to~82\%. These results underscore the need for higher-quality audio resources to bridge the gap with multimodal systems.

The EmoBox~\cite{ma2024emoboxmultilingualmulticorpusspeech} initiative attempts to address these challenges by aggregating multiple corpora into one repository, thereby easing the access but leaving the heterogeneity unresolved. Researchers must still download each dataset individually and navigate its unique characteristics, such as different sampling rates and metadata formats. Meanwhile, benchmarks such as EmoBench~\cite{sabour2024emobenchevaluatingemotionalintelligence} evaluate emotional intelligence in LLMs using hand-crafted questions. However, these benchmarks remain text-only and exclude audio.

Our work builds on these efforts by focusing on standardizing emotional speech datasets and enabling the multilingual evaluation of SER systems through text instructions and audio prompts.

\begin{figure*}[htbp]
    \centering
    \includegraphics[width=\textwidth]{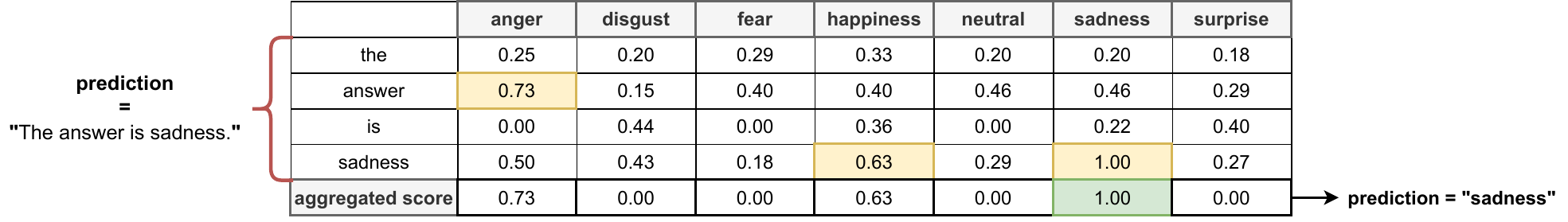}
    \caption{Example illustrating the application of the post-processing strategy used to extract labels from generated responses.}
    \label{fig:postprocessing-strategy}
\end{figure*}

\section{Corpus Design}
\label{sec:corpus-design}

The \textbf{CAMEO} collection was designed to provide accessible, multilingual emotional speech data and ensure the reproducibility of evaluation results. The goal was to compile relevant emotion recognition corpora that could also be used for other speech-related research. To support cross-lingual and cross-cultural studies, we prioritized linguistic diversity by including a broad range of languages.

All corpora included in the collection are freely available under licenses that permit non-commercial use and derivative works. Licensing information is included in the metadata. To maximize usability, the collection is accompanied by extensive documentation and standardized metadata. Additionally, open-source evaluation code is provided to enable benchmarking and facilitate reproducibility.

Since the datasets are public and can be used for model training, the collection does not define train or test splits. This decision reflects the intended use of the collection as a benchmark for evaluating cross-lingual model performance. Moreover, taking into consideration the limited amount of data usable for SER, further splitting would reduce diversity and not solve the problem of contamination.

\subsection{Selection Criteria for the Included Datasets}

The inclusion of a dataset in the collection was determined by the following criteria:
\begin{enumerate}
    \item \textbf{Availability and License:} The corpus is freely accessible and licensed for non-commercial use and derivative works.
    \item \textbf{Transcription:} The dataset contains transcriptions, either directly or through associated publications, to increase its usability for other speech-related tasks.
    \item \textbf{Emotions:} The annotations of basic emotional states are provided and consistent with standard taxonomies, such as Plutchik's Wheel.
    \item \textbf{Metadata (optional):} Information about speakers (IDs and demographics) was considered valuable. If such metadata was not available, relavant details were derived from documentation or publications.
\end{enumerate}

\subsection{Selected Corpora}
\label{sec:selected-corpora}

The \textbf{CAMEO} collection consists of 13 carefully selected datasets, which are presented in Table~\ref{tab:datasets}.

\begin{table}[htbp]
\centering
\begin{tabular}{lcc}
    \hline
    \textbf{Dataset} & \textbf{Language} & \textbf{No. samples} \\
    \hline
    CaFE \cite{cafe} & French & 936 \\
    CREMA-D \cite{cremad} & English & 7\,442 \\
    EMNS \cite{emns} & English & 1\,205 \\
    Emozionalmente \cite{emozionalmente} & Italian & 6\,902 \\
    eNTERFACE \cite{enterface} & English & 1\,257 \\
    JL-Corpus \cite{jlcorpus} & English & 2\,400 \\
    MESD \cite{mesd,mesd2} & Spanish & 862 \\
    nEMO \cite{christop2024nemodatasetemotionalspeech} & Polish & 4\,481 \\
    Or\'eau \cite{Oreau} & French & 502 \\
    PAVOQUE \cite{pavoque} & German & 5\,442 \\
    RAVDESS \cite{ravdess} & English & 1\,440 \\
    RESD \cite{resd} & Russian & 1\,396 \\
    SUBESCO \cite{subesco} & Bengali & 7\,000 \\ 
    \hline
\end{tabular}
\caption{Summary of the datasets included in the CAMEO collection.}
\label{tab:datasets}
\end{table}

The collection consists of a total of~41\,265~audio~samples. English accounts for approximately one-third of the material, reflecting its dominance in SER research. The collection represents~17 different emotions, with over~93\% of samples annotated with one of the seven primary states: anger, disgust, fear, happiness, neutrality, sadness, and surprise.

Speaker-related metadata is widely available. Specifically,~94.5\% of the samples include speaker identifiers, and~92.9\% include gender annotations. Of these samples, 43.6\% are spoken by female speakers, and 49.3\% are spoken by male speakers.

\subsection{Dataset Curation and Processing}

The development of the \textbf{CAMEO} collection followed a structured, multi-stage process:
\begin{enumerate}
    \item \textbf{Data Acquisition:} All datasets were downloaded from their original sources, inspected for consistency, assigned unique IDs, and purged of duplicates.
    \item \textbf{Standardization:} Audio samples were converted to FLAC (16-bit, 16~kHz). Metadata and transcriptions were unified across datasets. Whisper~Large~v2~\cite{whisper2022} was used to align transcriptions when necessary.
    \item \textbf{Metadata Serialization:} Metadata was stored in JSON Lines (UTF-8) format and linked to sample IDs.
    \item \textbf{Distribution:} The curated collection was uploaded to Hugging Face with proper credit given to the dataset authors. The metadata fields are summarized in Table~\ref{tab:metadata-fields}.
\end{enumerate}

\begin{table}[htbp]
\centering
\begin{tabular}{lp{0.68\columnwidth}}
    \hline
    \textbf{Field} & \textbf{Description} \\
    \hline
    file\_id        & Unique identifier of the audio sample. \\
    audio           & File path, raw waveform, and sampling rate. \\
    emotion         & Expressed emotional state. \\
    transcription   & Orthographic transcription of the utterance. \\
    speaker\_id     & Unique speaker identifier. \\
    gender          & Gender of the speaker. \\
    age             & Age of the speaker. \\
    dataset         & Dataset of origin. \\
    language        & Primary language of the sample. \\
    license         & Original dataset license. \\
    \hline
\end{tabular}
\caption{Overview of the metadata fields available in the CAMEO corpus, as published on the Hugging Face platform.}
\label{tab:metadata-fields}
\end{table}

\section{Evaluation}
\label{sec:evaluation}

The evaluation protocol was designed based on the following principles:
\begin{enumerate}
    \item To ensure reproducibility of the results, the evaluation includes easily accessible systems with audio modality.
    \item To promote transparency and allow comparison across systems, evaluation results are publicly available on a leaderboard hosted on the Hugging Face platform.
    \item To capture the performance of the models across different emotions and languages, the evaluation employs widely used metrics, such as macro-averaged F1 score, weighted F1 score and accuracy.
    \item To enable further development, the benchmark is designed to be easily extensible. It allows for the easy integration of new datasets, metrics, methods, and systems.
\end{enumerate}

To ensure consistency and fairness, all systems were evaluated using the same strategy. Each model was given a textual instruction and a single audio input. No additional metadata, such as speaker identity, gender, or language, was provided during inference. 

The models were expected to produce a single-word output corresponding to the predicted emotional state. However, due to the generative nature of the evaluated systems, the models occasionally generated descriptive responses despite the given instructions. Additionally, the models tended to respond with an adjective rather than a noun. 

As SER already poses a challenge, a post-processing strategy was designed to ensure that the models would not be penalized for minor errors. If a generated response is not an exact match for any of the labels, it is normalized and split into words. Then, the Levenshtein ratio between each target label and each word in the generated response is calculated. Similarity scores below a predefined threshold of~0.57 are filtered out for each label. The value of threshold was determined based on the Levenshtein ratio between the noun and adjective forms of the seven primary emotional states, as shown in Table~\ref{tab:threshold}. After filtering out the similarity scores, the remaining values are summed to yield an aggregated score for the given label. The label with the highest aggregated similarity score is selected as the best match. An example illustrating the application of the post-processing strategy is presented in Figure~\ref{fig:postprocessing-strategy}.

\begin{table}[htbp]
\centering
\begin{tabular}{ccc}
    \hline
    \textbf{Noun} & \textbf{Adjective} & \textbf{Levenshtein Ratio} \\
    \hline
    anger & angry & 0.8000 \\
    disgust & disgusted & 0.8750 \\
    fear & fearful & 0.7273 \\
    happiness & happy & 0.5714 \\
    neutrality & neutral & 0.8235 \\
    sadness & sad & 0.6000 \\
    surprise & surprised & 0.9412 \\
    \hline
\end{tabular}
\caption{Levenshtein similarity ratios (LR) between the noun and adjective forms of the seven most frequently occuring emotions in the selected datasets.}
\label{tab:threshold}
\end{table}

\begin{table*}[htbp]
\centering
\begin{tabular}{lccccccccc}
    \hline
    \textbf{Model} & \textbf{Bengali} & \textbf{English} & \textbf{French} & \textbf{German} & \textbf{Italian} & \textbf{Polish} & \textbf{Russian} & \textbf{Spanish} & \textbf{All} \\
    \hline
    \textbf{Ichigo} $_{0.7}$ & 0.1253 & 0.1293 & 0.1351 & 0.1226 & 0.1250 & 0.1489 & 0.1343 & 0.1634 & 0.1277 \\
    \textbf{Qwen2} $_{0.0}$ & \textbf{0.2508} & \textbf{0.3435} & \textbf{0.3154} & \textbf{0.3341} & \textbf{0.1913} & \textbf{0.1767} & \textbf{0.2450} & \textbf{0.2333} & \textbf{0.2023} \\
    \textbf{SeaLLMs} $_{0.3}$ & 0.0628 & 0.2168 & 0.1896 & 0.1216 & 0.0999 & 0.0547 & 0.0976 & 0.1761 & 0.1599 \\
    \textbf{Ultravox} $_{0.7}$ & 0.1163 & 0.1680 & 0.3073 & 0.2841 & 0.1383 & 0.1392 & 0.2086 & 0.2175 & 0.1484 \\
    \hline
    \textbf{Average} & 0.1388 & 0.2144 & 0.2369 & 0.2156 & 0.1386 & 0.1299 & 0.1714 & 0.1976 & 0.1596 \\
    \hline
\end{tabular}
\caption{F1 macro scores obtained by each model, at a selected temperature setting, reported per language.}
\label{tab:results-language}
\end{table*}

\begin{table*}[htbp]
\centering
\begin{tabular}{lccccccc}
    \hline
    \textbf{Model} & \textbf{anger} & \textbf{disgust} & \textbf{fear} & \textbf{happiness} & \textbf{neutral} & \textbf{sadness} & \textbf{surprise} \\
    \hline
    \textbf{Ichigo} $_{0.7}$ & 0.1414 & 0.1317 & 0.1366 & 0.1502 & 0.0377 & 0.2021 & 0.0793 \\
    \textbf{Qwen2} $_{0.0}$ & 0.1618 & 0.3211 & \textbf{0.1822} & 0.0367 & 0.0255 & \textbf{0.3905} & \textbf{0.4783} \\
    \textbf{SeaLLMs} $_{0.3}$ & 0.2246 & 0.2274 & 0.0153 & 0.0069 & 0.0135 & 0.2668 & 0.0956 \\
    \textbf{Ultravox} $_{0.7}$ & \textbf{0.2270} & \textbf{0.3420} & 0.0593 & \textbf{0.1673} & \textbf{0.0609} & 0.2193 & 0.1728 \\
    \hline
    \textbf{Average} & 0.1887 & 0.2556 & 0.0984 & 0.0903 & 0.0344 & \textbf{0.2697} & 0.2065 \\
    \hline
\end{tabular}
\caption{F1 scores obtained for primary emotional states across the full corpus.}
\label{tab:results-emotion}
\end{table*}

\section{Benchmark Results}
\label{sec:benchmark-results}

Four different models were evaluated: Qwen2-Audio-7B-Instruct~\cite{chu2024qwen2audiotechnicalreport}, Ichigo-llama3.1-s-instruct-v0.4~\cite{Ichigo}, SeaLLMs-Audio-7B~\cite{zhang2024seallms3openfoundation} and ultravox-v0\_5-llama-3\_1-8b~\cite{ultravox2024}. To examine their behavior under different conditions, each system was evaluated using three different temperature settings:~0.0, 0.3, and~0.7.


Table~\ref{tab:results-overall} reports the overall performance using macro-averaged F1 score, weighted F1 score, and accuracy. Performance remained low across all models and temperatures, highlighting the difficulty of multilingual speech emotion recognition in a zero-shot setting. The best-performing system was Qwen2-Audio, which achieved a macro F1 score of~0.20 at a temperature of~0.0. Though modest, this result is consistent with the absence of task-specific fine-tuning and emphasizes the difficulty of mapping acoustic emotion cues to specific labels across languages using general-purpose AudioLLMs.

\begin{table}[htbp]
\centering
\begin{tabular}{lccc}
    \hline
    \textbf{Model} & \textbf{Macro F1} & \textbf{Weighted F1} & \textbf{Accuracy} \\
    \hline
    Ichigo $_{0.0}$ & 0.1164 & 0.1171 & 0.1518 \\
    Ichigo $_{0.3}$ & 0.1164 & 0.1262 & \textbf{0.1549} \\
    Ichigo $_{0.7}$ & \textbf{0.1277} & \textbf{0.1359} & 0.1494 \\
    \hline
    Qwen2 $_{0.0}$ & \textbf{0.2023} & \textbf{0.3710} & \textbf{0.3911} \\
    Qwen2 $_{0.3}$ & 0.1985 & 0.3704 & 0.3387 \\
    Qwen2 $_{0.7}$  & 0.2005 & 0.3591 & 0.3825 \\
    \hline
    SeaLLMs $_{0.0}$ & 0.1461 & \textbf{0.1887} & \textbf{0.2383} \\
    SeaLLMs $_{0.3}$ & \textbf{0.1599} & 0.1782 & 0.2200 \\
    SeaLLMs $_{0.7}$ & 0.1399 & 0.1616 & 0.2135 \\
    \hline
    Ultravox $_{0.0}$ & 0.1417 & \textbf{0.2061} & \textbf{0.2333} \\
    Ultravox $_{0.3}$ & 0.1369 & 0.1972 & 0.2251 \\
    Ultravox $_{0.7}$ & \textbf{0.1484} & 0.2014 & 0.2294 \\
    \hline
\end{tabular}
\caption{Overall evaluation results computed across all samples in the corpus.}
\label{tab:results-overall}
\end{table}


Table~\ref{tab:results-language} shows the macro F1 scores for each language at a fixed temperature. Qwen2-Audio consistently outperformed the other models across all evaluated languages, suggesting stronger generalization capabilities or broader exposure to emotionally annotated speech during pretraining. Qwen2-Audio demonstrated notably higher performance for English, French, and German. Ultravox also demonstrated higher performance for French and German, albeit to a lesser extent. These results suggest the possibility of data contamination, especially since these languages are well represented in publicly available emotional speech corpora.



To investigate potential overlap effects, macro F1 scores were computed at the dataset level. Qwen2-Audio performed substantially better on CREMA-D~(0.80), RAVDESS~(0.73), and eNTERFACE~(0.54). A similar pattern was observed for Ultravox, which showed elevated performance on eNTERFACE~(0.79) and Or\'eau~(0.58). Since these datasets are widely used benchmarks in SER, the results suggest that they were likely encountered during training or fine-tuning.


Table~\ref{tab:results-emotion} shows F1 scores for the sever primary emotional states. Sadness was the most reliably recognized emotion across all systems, while neutral speech was the most challenging. This may be due to the fact that sadness often exhibits more stable and distinctive acoustic cues, such as reduced pitch variation and slower tempo. In contrast, neutral speech is inherently heterogeneous and often acoustically overlaps with low-arousal emotional states. Despite its narrative or conversational nature, the lack of salient prosodic markers likely contributes to frequent misclassification.

\section{Conclusions}
\label{sec:conclusions}

This work introduces \textbf{CAMEO}, a publicly available collection of multilingual emotional speech datasets designed to facilitate transparent, reproducible benchmarking for SER. The corpus is released alongside evaluation tools and an open leaderboard for model comparison. The evaluation of several systems with the audio modality revealed low overall performance, highlighting the difficulty of the SER task, especially in multilingual and zero-shot settings.

Future work will focus on several key areas. First, the \textbf{CAMEO} collection will be expanded to include additional datasets, particularly those representing low-resource languages and underrepresented emotional states. Second, we intend to evaluate new models to make \textbf{CAMEO} a continuously developing resource for advancing SER research.

\bibliographystyle{IEEEbib}
\bibliography{bibliography}

@misc{ma2024emoboxmultilingualmulticorpusspeech,
      title={{EmoBox: Multilingual Multi-corpus Speech Emotion Recognition Toolkit and Benchmark}}, 
      author={Z. Ma and M. Chen and H. Zhang and Z. Zheng and W. Chen and X. Li and J. Ye and X. Chen and T. Hain},
      year={2024},
      eprint={2406.07162},
      archivePrefix={arXiv},
      primaryClass={cs.SD},
      url={https://arxiv.org/abs/2406.07162}, 
}

@ARTICLE{10879457,
    author={Catania, F. and Wilke, J. W. and Garzotto, F.},
    journal={IEEE Transactions on Audio, Speech and Language Processing}, 
    title={{Emozionalmente: A Crowdsourced Corpus of Simulated Emotional Speech in Italian}}, 
    year={2025},
    volume={33},
    pages={1142-1155},
    doi={10.1109/TASLPRO.2025.3540662}
}

@article{article_1214312, 
    title={{CREMA-D: Improving Accuracy with BPSO-Based Feature Selection for Emotion Recognition Using Speech}},
    journal={Journal of Soft Computing and Artificial Intelligence}, 
    volume={3}, 
    pages={51–57}, 
    year={2022}, 
    DOI={10.55195/jscai.1214312}, 
    author={Donuk, K.}, 
    number={2}, 
    publisher={Mahmut DİRİK} 
}

@misc{sabour2024emobenchevaluatingemotionalintelligence,
    title={{EmoBench: Evaluating the Emotional Intelligence of Large Language Models}}, 
    author={S. Sabour and S. Liu and Z. Zhang and J. M. Liu and J. Zhou and A. S. Sunaryo and J. Li and T. M. C. Lee and R. Mihalcea and M. Huang},
    year={2024},
    eprint={2402.12071},
    archivePrefix={arXiv},
    primaryClass={cs.CL},
    url={https://arxiv.org/abs/2402.12071}, 
}

@article{MOCANU2023104676,
    title = {{Multimodal emotion recognition using cross modal audio-video fusion with attention and deep metric learning}},
    journal = {Image and Vision Computing},
    volume = {133},
    pages = {104676},
    year = {2023},
    issn = {0262-8856},
    doi = {https://doi.org/10.1016/j.imavis.2023.104676},
    url = {https://www.sciencedirect.com/science/article/pii/S0262885623000501},
    author = {B. Mocanu and R. Tapu and T. Zaharia},
}

@article{whisper2022,
    title = {{Robust Speech Recognition via Large-Scale Weak Supervision}},
    author = {A. Radford and J. W. Kim and T. Xu and G. Brockman and C. McLeavey and I. Sutskever},
    doi = {10.48550/ARXIV.2212.04356},
    year = {2022},
}

@inproceedings{cafe,
    author = {Gournay, P. and Lahaie, O. and Lefebvre, R.},
    title = {{A Canadian French Emotional Speech Dataset}},
    year = {2018},
    isbn = {9781450351928},
    publisher = {Association for Computing Machinery},
    address = {New York, NY, USA},
    url = {https://doi.org/10.1145/3204949.3208121},
    doi = {10.1145/3204949.3208121},
    booktitle = {Proceedings of the 9th ACM Multimedia Systems Conference},
    pages = {399–402},
    numpages = {4},
    location = {Amsterdam, Netherlands},
    series = {MMSys '18}
}

@article{cremad,
    author = {Cao, H. and Cooper, D. and Keutmann, M. and Gur, R. and Nenkova, A. and Verma, R.},
    year = {2014},
    pages = {377-390},
    title = {{CREMA-D: Crowd-sourced emotional multimodal actors dataset}},
    volume = {5},
    journal = {IEEE transactions on affective computing},
    doi = {10.1109/TAFFC.2014.2336244}
}

@misc{emns,
    title={{EMNS /Imz/ Corpus: An emotive single-speaker dataset for narrative storytelling in games, television and graphic novels}}, 
    author={K. A. Noriy and X. Yang and J. J. Zhang},
    year={2023},
    eprint={2305.13137},
    archivePrefix={arXiv},
    primaryClass={cs.CL},
    url={https://arxiv.org/abs/2305.13137}, 
}

@article{emozionalmente,
    author = {Catania, F. and Wilke, J. and Garzotto, F.},
    year = {2025},
    pages = {1-14},
    title = {{Emozionalmente: A Crowdsourced Corpus of Simulated Emotional Speech in Italian}},
    volume = {PP},
    journal = {IEEE Transactions on Audio, Speech and Language Processing},
    doi = {10.1109/TASLPRO.2025.3540662}
}

@INPROCEEDINGS{enterface,
    author={Martin, O. and Kotsia, I. and Macq, B. and Pitas, I.},
    booktitle={22nd International Conference on Data Engineering Workshops (ICDEW'06)}, 
    title={{The eNTERFACE' 05 Audio-Visual Emotion Database}}, 
    year={2006},
    pages={8-8},
    doi={10.1109/ICDEW.2006.145}
}

@inproceedings{jlcorpus,
    author = {James, J. and Tian, L. and Watson, C.},
    year = {2018},
    pages = {2768-2772},
    title = {{An Open Source Emotional Speech Corpus for Human Robot Interaction Applications}},
    doi = {10.21437/Interspeech.2018-1349}
}

@inproceedings{mesd,
    author = {Duville, M. M. and Alonso-Valerdi, L. and Ibarra-Zarate, D. I.},
    year = {2021},
    title = {{The Mexican Emotional Speech Database (MESD): elaboration and assessment based on machine learning}},
    volume = {2021},
    doi = {10.1109/EMBC46164.2021.9629934}
}

@article{mesd2,
    author = {Duville, M. M. and Alonso-Valerdi, L. and Ibarra-Zarate, D. I.},
    year = {2021},
    title = {{Mexican Emotional Speech Database Based on Semantic, Frequency, Familiarity, Concreteness, and Cultural Shaping of Affective Prosody}},
    volume = {6},
    journal = {Data},
    doi = {10.3390/data6120130}
}

@MISC{oreau,
    title = {{French emotional speech database - Or{\'e}au}},
    author = {Kerkeni, L. and Cleder, C. and Serrestou, Y. and Raoof, K.},
    publisher = {Zenodo},
    year      =  {2020}
}

@inproceedings{pavoque,
    author = {Steiner, I. and Schr\"oder, M. and Klepp, A.},
    title = {{The PAVOQUE corpus as a resource for analysis and synthesis of expressive speech}},
    booktitle = {Phonetik \& Phonologie 9. Phonetik \& Phonologie (P\&P-9), October 11-12, Zurich, Switzerland},
    year = {2013},
    pages = {83--84},
    organization = {UZH},
    publisher = {Peter Lang}
}

@article{ravdess,
    doi = {10.1371/journal.pone.0196391},
    author = {Livingstone, S. R. AND Russo, F. A.},
    journal = {PLOS ONE},
    publisher = {Public Library of Science},
    title = {{The Ryerson Audio-Visual Database of Emotional Speech and Song (RAVDESS): A dynamic, multimodal set of facial and vocal expressions in North American English}},
    year = {2018},
    volume = {13},
    url = {https://doi.org/10.1371/journal.pone.0196391},
    pages = {1-35},
    number = {5},
}

@misc{resd,
    author = {A. Amentes and N. Davidchuk and I. Lubenets},
    title = {{Russian Emotional Speech Dialogs with annotated text}},
    year = {2022},
    publisher = {Hugging Face},
    journal = {Hugging Face Hub},
    howpublished = {\url{https://huggingface.co/datasets/Aniemore/resd_annotated}},
}

@article{subesco,
    doi = {10.1371/journal.pone.0250173},
    author = {Sultana, S. AND Rahman, M. S. AND Selim, M. R. AND Iqbal, M. Z.},
    journal = {PLOS ONE},
    publisher = {Public Library of Science},
    title = {{SUST Bangla Emotional Speech Corpus (SUBESCO): An audio-only emotional speech corpus for Bangla}},
    year = {2021},
    volume = {16},
    url = {https://doi.org/10.1371/journal.pone.0250173},
    pages = {1-27},
    number = {4},
}

@misc{christop2024nemodatasetemotionalspeech,
    title={{nEMO: Dataset of Emotional Speech in Polish}}, 
    author={I. Christop},
    year={2024},
    eprint={2404.06292},
    archivePrefix={arXiv},
    primaryClass={cs.CL},
    url={https://arxiv.org/abs/2404.06292}, 
}

@misc{chu2024qwen2audiotechnicalreport,
    title={{Qwen2-Audio Technical Report}}, 
    author={Y. Chu and J. Xu and Q. Yang and H. Wei and X. Wei and Z. Guo and Y. Leng and Y. Lv and J. He and J. Lin and C. Zhou and J. Zhou},
    year={2024},
    eprint={2407.10759},
    archivePrefix={arXiv},
    primaryClass={eess.AS},
    url={https://arxiv.org/abs/2407.10759}, 
}

@article{Ichigo,
    title={{Llama3-S}},
    author={Homebrew Research},
    year={2024},
    url={https://huggingface.co/homebrewltd/llama3.1-s-2024-08-20}
}

@misc{zhang2024seallms3openfoundation,
    title={{SeaLLMs 3: Open Foundation and Chat Multilingual Large Language Models for Southeast Asian Languages}}, 
    author={W. Zhang and H. P. Chan and Y. Zhao and M. Aljunied and J. Wang and C. Liu and Y. Deng and Z. Hu and W. Xu and Y. K. Chia and X. Li and L. Bing},
    year={2024},
    eprint={2407.19672},
    archivePrefix={arXiv},
    primaryClass={cs.CL},
    url={https://arxiv.org/abs/2407.19672}, 
}

@misc{ultravox2024,
    author       = {{Fixie AI}},
    title        = {{Ultravox v0.5 (Llama 3.1 8B)}},
    year         = {2024},
    howpublished = {\url{https://huggingface.co/fixie-ai/ultravox-v0_5-llama-3_1-8b}},
}

\end{document}